\documentclass[ukenglish]{nik}

\usepackage{graphicx}
\usepackage{hyperref}

\title{DIT4BEARs Smart Roads Internship 
}
\author{Md Abrar Jahin\\
\small Department of Industrial Engineering and Management\\
\small Khulna University of Engineering and Technology (KUET), Bangladesh\\
\small \href{mailto:abrar.jahin.2652@gmail.com}{abrar.jahin.2652@gmail.com}\\
\and
Andrii Krutsylo\\
\small Doctoral School of Information and Biomedical Technologies\\
\small Institute of Computer Science Polish Academy of Sciences\\
\small \href{mailto:andrii.krutsylo@protonmail.com}{andrii.krutsylo@protonmail.com}\\
}
\date{May 3, 2021 - May 21, 2021
}

\begin{document}

  \maketitle

  \begin{abstract}
The research internship at UiT - The Arctic University of Norway was offered for our team being the winner of the 'Smart Roads - Winter Road Maintenance 2021' Hackathon. The internship commenced on 3 May 2021 and ended on 21 May 2021 with meetings happening twice each week. In spite of having different nationalities and educational backgrounds, we both interns tried to collaborate as a team as much as possible. The most alluring part was working on this project made us realize the critical conditions faced by the arctic people, where it was hard to gain such a unique experience from our residence. We developed and implemented several deep learning models to classify the states (dry, moist, wet, icy, snowy, slushy). Depending upon the best model, the weather forecast app will predict the state taking the Ta, Tsurf, Height, Speed, Water, etc. into consideration. The crucial part was to define a safety metric which is the product of the accident rates based on friction and the accident rates based on states. We developed a regressor that will predict the safety metric depending upon the state obtained from the classifier and the friction obtained from the sensor data. A pathfinding algorithm has been designed using the sensor data, open street map data, weather data.

  \end{abstract}

  \section{Internship Description}
Monitoring of road conditions, especially in the Barents-Euro Arctic region, can be very helpful to different
parties. For example, the city could benefit from road condition information for different tasks such as
salting, getting rid of snow, and planning of public transportation routes.
An RCM411 sensor [1] has been installed in a waste collection vehicle operating in the city of Narvik as a
part of the DIT4BEARs project. It is an optical sensor that collects measurements such as the coefficient of
friction and water layer thickness. The measurements are sent to a data storage.
The task of this internship is to use the sensor measurements and any other publicly available data, e.g.,
meteograms [3], to build a deep learning model that can predict the roads state. Then, apply this model to
design a path planning algorithm that helps a driver to plan the safest route possible given the expected
conditions on the roads that constitute that route, while considering topological maps [2].

\subsection{Research Analysis}
\begin{itemize}
    \item Data was collected and processed 
    \item Designed and implemented a deep learning model that identifies the road state as one of the following
states: Dry, Moist, Wet, Icy, Snowy, Slushy. One possible application is to predict the need for
salting and the required amount of salt.
    \item Designed a path planning algorithm for a city like Narvik based on the trained model, while taking
the topography into account.
\end{itemize} 

\subsection{Technologies Used}
Python 3, sklearn models (Elastic Net, Ridge, Lasso, Linear Regression, K-Neighbors Classifier), XGBClassifier, XGBRegressor, Prophet, pandas, NumPy, Jupyter Notebook, Github, Git bash, OSMNX, PyOWM and WWO$\_$hist

\section{Overview of internship experience}
During our internship experience with UiT - The Arctic University of Norway, we were able to develop our communication skills (cross-cultural collaboration, leadership, stress management, teamwork, adaptability, critical thinking, interpersonal communication, friendly personality). We particularly found Machine Learning research experience to be useful in improving our technical skills (Data Analysis and Visualization using Python machine learning modules). Although we encountered working with the real-world data to be challenging, we found it to be valuable in developing our statistical analysis, diagnostic analysis, inferential analysis, predictive analysis - forecasting skills. While data analysis process, we gradually improved data cleaning, analysis, interpretation, visualization techniques by testing different methods to fit our model with the real-world data.

\section{Data}
In this section, description of input data and data flow are as follows:\\

{\sl \textbf{Sensors}} - Data collected from RCM411 sensors, merged into one dataframe. Number of records was reduced by rounding latitude and longitude from 6 decimal place to 4, what gives the accuracy of 36 feet. Next parameters was selected:
 \begin{itemize}
    \item Friction: Measured friction value between 0.1-0.81
    \item State: Values of 1-6 for Dry, Moist, Wet, Icy, Snowy, Slushy
    \item Ta: Air temperature
    \item Tsurf: Surface temperature
    \item Water: Water layer thickness in 0-3 mm
    \item Speed: Wind speed
    \item Latitude
    \item Longitude
    \item Height: Above sea-level in m
    \item Accuracy \\
 \end{itemize}

{\sl \textbf{Open Steet Map}} - Data generated by OSMNX library from latitude and longitude of each record of sensors data. Next parameters was selected:
 \begin{itemize}
    \item osmid
    \item highway
    \item oneway
    \item length
    \item speed$\_$kph
    \item travel$\_$time
    \item bridge
    \item junction
    \item tunnel
    \item lanes \\
 \end{itemize}
 
{\sl \textbf{Weather}} - Data generated by PyOWM and WWO$\_$hist libraries from date, time, latitude and longitude of each record of sensors data. Next parameters was selected:
 \begin{itemize}
    \item maxtempC
    \item mintempC
    \item totalSnow$\_$cm
    \item sunHour
    \item uvIndex
    \item DewPointC
    \item HeatIndexC
    \item WindGustKmph
    \item cloudcover
    \item humidity
    \item precipMM
    \item pressure
    \item tempC
    \item visibility
    \item winddirDegree
    \item windspeedKmph
    \item UV
 \end{itemize}
 
  \subsection{Limitations}
 
 The data were collected in such a way that every day of observations contains only a small part of all unique coordinates visited over the entire period of observations. Moreover, daily observations are a series of records with constantly changing time and coordinates, which gives only one data point for specific location and time per day. For the entire observation period, the number of data points per unique coordinates is between 1 and 32 with average 2.93. 
 
 Available data does not allow to build a model that is capable to make predictions about specific location and time. 
 
\section{Solutions Developed}
\begin{enumerate}
    
\item Utilities for processing data from sensors and resources providing historical weather data.
\item Graph Neural Network ready for corrected collected in such a way as to have observations over the entire period from a variety of locations (for example, using the sensor network). 
\item The classifier for predicting one the six weather states was developed.
\item New safety metric was defined (safety = the accident rate for friction * the accident rate for weather states).
\item The regressor for forecasting the defined safety metric of the road was developed.

\end{enumerate}

\section{Recommendations and Updates }
While we had many useful experiences at the UiT - The Arctic University of Norway DIT4BEARs Research project, we feel that we still need to develop our confidence levels with real-world data preprocessing and machine learning techniques[tasks]. I would have enjoyed more time completing the classification (target variable: State) and regression (target variable: Safety metric) using more sophisticated models. We were trying different methods of preprocessing such as: removing duplicates of the most crucial features. There were adequacy of necessary data such as the data might have been collected in each hour or each week instead of each minute. This drawback frustrated us a lot while data cleaning process. 

\section{Literature Review}
The data of accident rates for different weather states and friction coefficients have been collected from \textbf{Friction measurement methods
and the correlation between road
friction and traffic safety}. The authors of Friction and Traffic Safety project aimed to show the quantitative correlation between friction and road accident risks. Also they collected information from various friction measurement methods. They maintained that drivers behavior is adjusted depending upon the road friction, tyre sound, sliding and skidding movement of vehicles, weather states. When temperature decreases, the friction coefficient decreases and the accident risk increases dramatically whereas there is no sufficient and reliable amount of correlation between accident risk and friction. However, for the wet or dry road, the road friction remains reasonable and accident risk also remains within safety threshold. The surface temperature of icy road affects the friction to a great extent. For instance, when the temperature rises, the ice starts to melt and the road becomes slippery resulting the accident risk to be higher [4].\\
The optical sensors and the local weather forecasts is being effectively used for road condition forecast and collection of slippery road information. Machine learning models for the traffic flow information, weather state forecast, information about salt deployed on the road is the place of improvement in case of road condition forecast [5].

\section{Source code}
Open access: \url{https://github.com/DentonJC/DIT}

\section*{Acknowledgements}
This internship report summarises DIT4BEARS Smart Roads Project that made possible through the kind mentorship and coordination of following personalities-
\begin{itemize}
    \item Mentor: Per Arne Sundsbø
    \item Coordinators: Ghada Bouzidi, Aleksander Pedersen, Rune Dalmo
\end{itemize}
to whom the authors would like to express their gratitude. The authors are greatly indebted to them for their valuable comments on this report.\\
\textbf{Place of Internship:} UiT – The Arctic University of Norway, Faculty of Engineering Science and Technology, Department of Computer Science and Computational Engineering

  \bibliographystyle{plain}
  \bibliography{bibliography}

\begin{thebibliography}{9}
\bibitem{teconer} \url{https://www.teconer.fi/en/surface-condition-friction-measurements/#RCM411}
\bibitem{openstreetmap} \url{https://wiki.openstreetmap.org/wiki/AboutOpenStreetMap}
\bibitem{yr} \url{https://www.yr.no}
\bibitem{accident_rates}
Carl-Gustaf Wallman \& Henrik Åström: \textbf{Friction measurement methods
and the correlation between road
friction and traffic safety}, VTI meddelande 911A. 2001
\bibitem{intelligent road}
J. Autioniemi, M. Autioniemi, J. Casselgren, H. Konttaniemi, T. Sukuvaara , R. Ylitalo: \textbf{Intelligent Road}, Lapland University of Applied Sciences, Luleå University of Technology, Finnish Meteorological Institute

\end{thebibliography}

\end{document}